# Parametrical Neural Networks and Some Other Similar Architectures

Leonid B. Litinskii

Institute of Optical Neural Technologies Russian Academy of Sciences,
Moscow, litin@mail.ru

## Abstract

A review of works on associative neural networks accomplished during last four years in the Institute of Optical Neural Technologies RAS is given. The presentation is based on description of parametrical neural networks (PNN). For today PNN have record recognizing characteristics (storage capacity, noise immunity and speed of operation). Presentation of basic ideas and principles is accentuated.

## 1. Introduction

The Hopfield Model (HM) is a well-known version of a binary auto-associative neural network. With the aid of this model one can retrieve binary $N$-dimensional input patterns when their distorted copies are given. As is well known, the storage capacity of HM is rather small. The statistical physics approach gives the estimate $M_{HM} \sim 0.1 \cdot N$, where $M_{HM}$ is a number of randomized patterns, which can be handled by HM [1]. At the same time, elementary calculations show that if the number of patterns is less than $N$, the most effective approach is to store patterns in the computer directly and to associate the input vector with that pattern, the distance to which is the smallest one. Thus, the small storage capacity of HM makes it useless for practical applications.

In the end of 80-th attempts were made to improve recognizing characteristics of auto-associative memory by considering $q$-nary patterns, whose coordinates can take $q>2$ different values ([2]-[5]). These models were designed for handling of color images. The number $q$ in this case is the number of different colors, $q \sim 10^2 - 10^3$. All these models (with one exception) had the storage capacity even smaller than HM. The exception was the Potts-glass neural network (PG), for which the statistical physics approach gave the estimate $M_{PG} \sim \frac{q(q-1)}{2} M_{HM}$ [3]. Since in this case the storage capacity is rather large, $M_{PG} \sim 10^4 \cdot N$, the recognizing characteristics of Potts-glass neural network are of interest for practical applications. Unfortunately, the reasons of such large storage capacity of Potts-glass neural network were not understood. The statistical physics approach does not provide an explanation for this result.

On the other hand, in [6],[7] the optical model of auto-associative memory was examined. Such a network is capable to hold and handle information that is encoded in the form of the phase-frequency modulation. In the network the signals propagate along interconnections in the form of quasi-monochromatic pulses at $q$ different frequencies. The model is based on *a parametrical neuron* that is a cubic nonlinear element capable to transform and generate pulses in four-wave mixing processes [8]. If the interconnections have a generalized Hebbian form, the storage capacity of such a network also exceeds $M_{HM}$ in $\frac{q(q-1)}{2}$ times. The authors of [6],[7] called their model *the parametrical neural network* (PNN).

Further analysis of PNN showed that it can be described adequately in the framework of the vector formalism, when $q$ different states of neurons are described by basis vectors of $q$-dimensional space. Everything else in this model (interconnections, a dynamical rule) is a direct generalization of HM.

The vector formalism allowed us to determine the identity of basic principles account for Potts-glass neural network and PNN. We succeeded in the understanding of the mechanism of suppression of internal noise, which guarantees high recognizing properties for both architectures. Some variants of auto-associative PNN-architectures were suggested, including the one that has the record storage capacity for today (PNN2). Moreover, we managed to build up: 1) *hetero-associative* variant of a $q$-nary neural network that in order of magnitude exceeds the auto-associative PNN with respect to the speed of operations; 2) *decorrelating PNN* that is an architecture aimed at handling of binary patterns; it allows one to store and recognize as much patterns as $\sim N^c$, where $c \approx 2-4$.

In the present review we summarize the results obtained in [9]-[22] by a group of authors from Institute of Optical Neural Technologies Russian Academy of Sciences.

## 2. Vector formalism for $q$-nary networks

### 2.1. PNN-architecture

To describe $q$ different states of neurons we will use unit vectors $\mathbf{e}_l$ of the space $\mathbf{R}^q$, $q \geq 1$:

$$\mathbf{e}_l = \begin{pmatrix} 0 \\ \vdots \\ 1 \\ \vdots \\ 0 \end{pmatrix}, \; l = 1, \ldots, q. \qquad (1)$$

The state of the $i$th neuron is given by the vector $\mathbf{x}_i$,

$$\mathbf{x}_i = x_i \mathbf{e}_{l_i}, \; x_i = \pm 1, \; \mathbf{e}_{l_i} \in \mathbf{R}^q, \; 1 \le l_i \le q, \; i = 1, \ldots, N.$$

If we compare vector formalism with the original optical model [6],[7], binary variable $x_i$ is modeling the presence of a phase $\phi = \{0/\pi\}$ of a quasi-monochromatic pulse, and unit vectors $\mathbf{e}_l$ are modeling the presence of $q$ different frequencies $\omega_l$.

As a whole $N$-dimensional image with $q$-nary coordinates is given as a set of $N$ $q$-dimensional vectors $\mathbf{x}_i$: $X = (\mathbf{x}_1, \ldots, \mathbf{x}_N)$, and $M$ initial patterns are $M$ given in advance similar sets:

$$\begin{aligned} X^{(\mu)} &= \left( \mathbf{x}_1^{(\mu)}, \mathbf{x}_2^{(\mu)}, \ldots, \mathbf{x}_N^{(\mu)} \right), \; \mathbf{x}_i^{(\mu)} = x_i^{(\mu)} \mathbf{e}_{l_i^{(\mu)}}, \\ x_i^{(\mu)} &= \pm 1, \; 1 \le l_i^{(\mu)} \le q, \; \mu = 1, \ldots, M. \end{aligned} \qquad (2)$$

As usual the local field at the $i$th neuron has the form

$$\mathbf{h}_i = \frac{1}{N} \sum_{j=1}^{N} \mathbf{T}_{ij} \mathbf{x}_j,$$

where $(q \times q)$-matrix $\mathbf{T}_{ij}$ defines interconnection between $i$th and $j$th neurons. The interconnections are chosen in the generalized Hebbian form:

$$\mathbf{T}_{ij} = \left( 1 - \delta_{ij} \right) \sum_{\mu=1}^{M} \mathbf{x}_i^{(\mu)} \mathbf{x}_j^{(\mu)+}, \; i, j = 1, \ldots, N, \qquad (3)$$

where $\mathbf{x}^+$ denotes $q$-dimensional vector-row, and $\mathbf{xy}^+$ denotes the product of vector-column $\mathbf{x}$ and vector-row $\mathbf{y}^+$ carried out according the matrix product rules. In other words, $\mathbf{xy}^+$ is a tensor product of two $q$-dimensional vectors. The fact that interconnections $\mathbf{T}_{ij}$ are defined by superposition of the states of $i$th and $j$th neurons over all patterns only, allows us to interpret them as Hebbian-like.

In the expression for the local field $\mathbf{h}_i$ the matrices $\mathbf{T}_{ij}$ act on the vectors $\mathbf{x}_j \in \mathbf{R}^q$. After summation over all $j$, the local field at the $i$th neuron will be a linear combination of unit vectors $\mathbf{e}_l$. The dynamic rule generalizing the standard asynchronous dynamics is defined as follows: at the moment $t+1$ the $i$th neuron is oriented in the direction that is the nearest to the direction of the local field $\mathbf{h}_i$ at the moment $t$. In other words, if at the moment $t$

$$\mathbf{h}_i(t) = \sum_{l=1}^{q} A_l^{(i)} \mathbf{e}_l, \quad \text{where} \quad A_l^{(i)} \sim \sum_{j(\neq i)}^{N} \sum_{\mu=1}^{M} \left(\mathbf{e}_l \mathbf{x}_i^{(\mu)}\right)\left(\mathbf{x}_j^{(\mu)} \mathbf{x}_j(t)\right),$$

and $A_k^{(i)}$ is the largest in modulus amplitude, $\left|A_k^{(i)}\right| = \max_{1 \leq l \leq q}\left|A_l^{(i)}\right|$, then

$$\mathbf{x}_i(t+1) = \text{sign}\left(A_k^{(i)}\right)\mathbf{e}_k.$$

The evolution of the system consists in successive orientation of vector-neurons according this rule. It is not difficult to show that during the process of evolution the energy of the state, $E(t) \sim -\sum_{i=1}^{N} \mathbf{x}_i^+(t)\mathbf{h}_i(t)$, decreases monotonically. Finally the system falls down to the local energy minimum that is a fixed point of the network.

This model was called PNN2 [9]-[12]. When $q=1$ it transforms into the standard HM.

## 2.2. Recognizing characteristics of PNN2

The estimates of recognizing characteristics of PNN2 are given in [9]-[12]. They can be obtained with the aid of the standard probability-theoretical approach [1], dividing the local field $\mathbf{h}_i$ into two parts, which are *the useful signal* and *the internal noise*, and calculating their average values and dispersions. In contrast to the standard situation described in [1], here partial noise components are not independent random values, but uncorrelated values only. This does not allow us to use the central limit theorem to estimate the characteristics of internal noise. However, one can use the Chebyshev-Chernov statistical method [22], which allows us to obtain the exponential part of the estimated value in the case of uncorrelated random variables also.

All calculations were performed accurately in [11],[12]. It was found that comparing with HM the dispersion of the internal noise is $q^2$ times less. This is because overwhelming majority of the values $\mathbf{x}_i^{(\mu)}\mathbf{x}_j^{(\mu)+}\mathbf{x}_j$, arising when calculating partial noisy components, vanishes. At that the dispersion of the internal noise decreases. While for HM the arising products $x_i^{(\mu)}x_j^{(\mu)}x_j$ can be "+1" or "-1" only. Consequently, each of these summands contributes to increase of the dispersion of the internal noise.

Let us right down the final expression for the probability of error of recognizing the input image, which is a distorted copy of one of the patterns. Let $a$ be the probability of distortion of coordinates of the pattern in spin variable $x_i$, and let $b$ be the probability of distortion of its coordinates in one of possible directions $\mathbf{e}_l$. (In terms of the original optical model [6],[7] $a$ is the probability of a distortion of the quasi-monochromatic pulse in its phase, and $b$ is the probability of a distortion in its frequency.) We suggested that the patterns (2) are randomized, e.g. their coordinates are independently equiprobable distributed. Then, we obtain the estimate for the retrieval error probability of the pattern in one-step retrieval, when the output pattern is evaluated from the input vector after one synchronous parallel calculation of all neurons:

$$\Pr_{err} < \sqrt{NM} \exp\left(-\frac{N(1-2a)^2}{2M} \cdot q^2 (1-b)^2\right). \qquad (4)$$

When $M, N \to \infty$ this probability tends to zero, if the number of patterns $M$ does not exceeds the critical value

$$M_{PNN2} = \frac{N(1-2a)^2}{2\ln N} \cdot q^2 (1-b)^2. \qquad (5)$$

The last quantity can be considered as asymptotically obtainable storage capacity of PNN2.

When $q = 1$ these expressions turn into the known results for HM (in this case there is no frequency noise and it follows that we have to set $b = 0$). When $q$ increases the retrieval error probability falls down exponentially: The noise immunity of the network increases significantly. In the same time the storage capacity (5) increases proportionally $q^2$. If we have in mind colored images processing, the neurons are pixels of the screen, and $q$ is the number of different colors. It can be considered that $q \sim 10^2 - 10^3$. With such $q$ the storage capacity of PNN2 by 4-6 orders of magnitude greater than the storage capacity of HM.

PNN2 allows one to store the number of patterns that is many times exceeds the number of neurons $N$. For example, let us set a constant value $\Pr_{err} = 0.01$. In the Hopfield model, with this retrieval error probability we can store $M = N/10$ patterns only, each of which is less then 30% noisy. In the same time, PNN2 with $q = 64$ allows us to retrieve any of $M = 5N$ patterns with 90% noise, or any of $M = 50N$ patterns with 65% noise. Computer simulations confirm these estimates.

### 2.3. PNN3

From the point of view of realization of PNN in the form of electronic device it is of interest its variant, when there are lacking of phases of quasi-monochromatic pulses (by which parametrical neurons exchange). If we use the language of the vector formalism, this means that to describe different states of neurons one use unit vectors (1) only, and the spin variable $x_i$ is absent:

$$\mathbf{x}_i = \mathbf{e}_{l_i}, \ \mathbf{e}_{l_i} \in \mathbf{R}^q, \ 1 \le l_i \le q, i = 1,\ldots,N. \qquad (6)$$

Now neurons can be found in one of $q$ different states, while for PNN2 the number of different states of neurons is equal to $2q$ (due to the spin variable $x_i$).

This model was called PNN3. It was analyzed in [12]-[14]. It was found that if as before the interconnections matrices $\mathbf{T}_{ij}$ are chosen in the form (3), the partial noise components are correlated. This leads to a disastrous increase of dispersion of internal noise. The way allowing us to overcome this difficulty is analogous to the one that is used in the sparse coding [23],[24]: when calculating the matrices $\mathbf{T}_{ij}$, it is necessary to subtract the value of the averaged neuron activity from the vector coordinates $\mathbf{x}_i^{(\mu)}$. For randomized patterns the averaged neuron activity is equal to $\mathbf{e}/q$, where $\mathbf{e}$ is the sum of all the unit vectors $\mathbf{e}_l$: $\mathbf{e} = \sum_1^q \mathbf{e}_l$. So, if in place of (3) we use

$$\mathbf{T}_{ij} = (1-\delta_{ij}) \sum_{\mu=1}^M (\mathbf{x}_i^{(\mu)} - \mathbf{e}/q)(\mathbf{x}_j^{(\mu)} - \mathbf{e}/q)^+, \ i,j = 1,\ldots,N, \qquad (7)$$

then partial noise components become uncorrelated, and we can use the Chebyshev-Chernov method. Here the analogues of the expressions (4), (5) are

$$\Pr_{err} < \sqrt{NM} \exp\left(-\frac{N}{2M} \cdot \frac{q(q-1)}{2}(1-\bar{b})^2\right). \qquad (8)$$

and

$$M_{PNN3} = \frac{N}{2\ln N} \cdot \frac{q(q-1)}{2}(1-\bar{b})^2, \ \text{where} \ \bar{b} = \frac{q}{q-1}b. \qquad (9)$$

The storage capacity of PNN3 is two times less comparing with PNN2. This is because in PNN3 the number of the states of neurons is two times less than in PNN2. On the whole these two models are very close with regard to their characteristics.

**Note.** PNN3 is a refined from useless difficulties version of the Potts-glass neural network, that was examined for the first time in [3]. For this publication poorly chosen notations and a brief style built up a reputation of a difficult and obscure text. However, in fact the Potts-glass neural network is almost coincides with PNN3. In both models interconnection matrices $\mathbf{T}_{ij}$ are defined with the aid of *the Potts vectors*, $\mathbf{x}_l = \mathbf{e}_l - \mathbf{e}/q$, which in the Potts-glass model are used to describe

$q$ different states of neurons; in PNN3 for this purpose one use unit vectors $\mathbf{e}_l$. Recognizing characteristics of PNN3 and Potts-glass neural network are identical.

In conclusion of this item we note again that recognizing characteristics of PNN-architecture improve when the number of states of neurons $q$ increases. We actively use this property in the next item.

## 3. Other architectures based on PNN

Outstanding recognizing characteristics of PNN can be used to develop other architectures, directed onto solution of one or another special problems. *The decorrelating PNN* allows one to store a polynomial large number of binary patterns, and, this is most important, even if these patterns are strongly correlated. *The q-nary identifier* allows one to accelerate the recognizing process by orders of magnitude, and in the same time to decrease computer memory necessary to store the interconnections.

### 3.1. The decorrelating PNN

It is known that the memory capacity of HM falls down drastically if there are correlations between patterns. The way out is so named sparse coding [23],[24]. The decorrelating PNN (DPNN) is an alternative to this approach [15]-[18].

The main idea of DPNN is as follows. The binary patterns are one-to-one mapped into internal representation using vector-neurons of large dimension $q$. Then PNN2 is being constructed on the basis of the obtained vector-neuron patterns. The mapping has the following properties: First, correlations between vector-neuron patterns become negligible, and, second, the dimension $q$ of vector-neurons increases exponentially as a function of the mapping parameter. The algorithm of binary patterns recognition consists of three stages. At first, the input binary vector is mapped into the $q$-nary representation. Then with the aid of PNN2 its recognition occurs: it is identified with one of the $q$-nary patterns. At last, the inverse mapping of the $q$-nary pattern into the binary representation takes place. Since the larger the dimension of vector-neurons, the better recognizing properties of PNN2, we can expect substantial increase of the storage capacity of the network.

The algorithm of binary patterns mapping into vector-neuron ones is very simple. Let $Y = (y_1, y_2, \ldots, y_N)$ be $N$-dimensional binary vector, $y_i = \pm 1$. We divide it into $n$ fragments of $k+1$ elements each, $N = n(k + 1)$. With each fragment we associate an integer number $l$ according the following rules: 1) the first element of the fragment defines the sign of the number; 2) the other $k$ elements of the fragment determine the absolute value of $l$,

$$l = 1 + \sum_{i=2}^{k+1}(y_i + 1) \cdot 2^{k-i}, \quad 1 \le l \le 2^k.$$

(In fact, we interpret the last $k$ elements of the fragment as the binary notation of the integer $l$.) After that we associate each fragment $(y_1, y_2, \ldots, y_{k+1})$ with a vector $\mathbf{x} = \pm \mathbf{e}_l$, where $\mathbf{e}_l$ is the $l$th unit vector in the space $\mathbf{R}^q$ of dimensionality $q = 2^k$. The sign of the vector is given by the first element of the fragment.

So, the mapping of binary fragment into $2^k$-dimensional unit vector occurs as follows:

$$(y_1, y_2, \ldots, y_{k+1}) \to \pm l \to \mathbf{x} = \pm \mathbf{e}_l \in \mathbf{R}^q, \quad q = 2^k. \qquad (10)$$

According this scheme any binary vector $Y \in \mathbf{R}^N$ can be one-to-one mapped into a set of $n$ $q$-dimensional unit vectors

$$Y = (y_1, \ldots, y_{k+1}, \ldots, y_{N-k}, \ldots, y_N) \to X = (\mathbf{x}_1, \mathbf{x}_2, \ldots, \mathbf{x}_n).$$

The result of this mapping is called *the internal image* of the binary vector $Y$. The number $k$ is called *the mapping parameter.*

When the mapping for the given set of binary patterns $\{Y^{(\mu)}\}_1^M$ is done, we obtain vector-neuron images $\{X^{(\mu)}\}_1^M$. Note, even if binary patterns $Y^{(\mu)}$ are correlated, their vector-neuron images are practically not correlated. Such is the property of the mapping (10): It eliminates the correlations between patterns. Indeed, for two binary fragments it is sufficient to differ even in one coordinate, and the fragments are mapped into two absolutely different orthogonal to each other unit vectors of the space $\mathbf{R}^q$.

Now, making use of the vector-neuron images $\{X^{(\mu)}\}_1^M$ we construct PNN2 – see formula (3). Since the images $X^{(\mu)}$ can be considered to be randomized, we use estimates (4),(5) obtained previously. If $a$ is the probability of distortion of binary coordinates (the level of noise), for constructed PNN2 we obtain:

$$M_{DPNN} \sim \frac{N(1-2a)^2}{2\ln N} \cdot \frac{(2(1-a))^{2k}}{k(1+k/\ln N)}. \qquad (11)$$

The first co-factor in the right-hand side is the storage capacity of HM. The second co-factor provides exponential in the mapping parameter $k$ increase of the storage capacity, since always the inequality $2(1-a) > 1$ is fulfilled.

However, it is necessary to be careful: the parameter $k$ cannot be an arbitrary large. First, the number of vector-neurons $n$ must be sufficiently large to make it possible to use probability-theoretic estimates. Experiments show that as far as the dimension of vector-neuron patterns is greater than 100, $n \geq 100$, asymptotical estimates (4),(5) are fulfilled to a high accuracy. This leads to the first restriction on the value of $k$:

$$k+1 \leq \frac{N}{100}. \qquad (12)$$

Second, among the vector-coordinates of the distorted vector-neuron image at least one coordinate must be undistorted. Its contribution to the local field $\mathbf{h}_i$ allows one to retrieve the correct value of the $i$th vector-coordinate. From these argumentations it follows that correct recognition can be guaranteed only if the number of undistorted vector-coordinates is not less than two. It is easy to estimate the number of undistorted vector-coordinates of the internal image X: $n(1-a)^{k+1}$. So, we obtain the second restriction on the value of $k$:

$$n(1-a)^{k+1} \geq 2. \qquad (13)$$

Thus, maximal possible value $k_c$ must fulfill the inequalities (12),(13).

When $k$ changes from 0 to $k_c$, the storage capacity of the network increases rapidly. When $k$ becomes greater than $k_c$, the network no longer retrieves the patterns. However, even before $k$ reaches the limiting value $k_c$, the storage capacity of the network can be done sufficiently large. Let us write down the storage capacity of DPNN (11) for the critical value $k_c$ in the form

$$M_{DPNN}(k_c) \sim N^R. \qquad (14)$$

In Fig.1 the dependences of $k_c$ (the top panel) and the exponent $R$ in the expression (14) (the bottom panel) on the level of the binary noise $a$ are shown for the dimension of the binary patterns $N$=1000. We see (the top panel) that for the most part of the variation interval of the noise $a$, the principal restriction for $k_c$ is inequality (12). This is because the dimension of binary patterns is comparatively small. Apparently, the reasonable level of the noise is $a$<0.2. The bottom panel shows that DPNN allows us to store $\sim N^2$ patterns. This is in orders of magnitude greater than the storage capacity of HM.

In Fig.2 the analogous graphs are shown for the dimension of binary patterns $N$=10000. Now the principle restriction for $k_c$ is the inequality (13). From the graph on the bottom panel it follows that for this dimension $N$ and the level of the noise $a$=0.1 the network can store $\sim N^6$ of binary patterns, and for $a$=0.15 we obtain $M_{DPNN} \sim N^4$. Note, these results are valid for any level of correlations between the binary patterns.

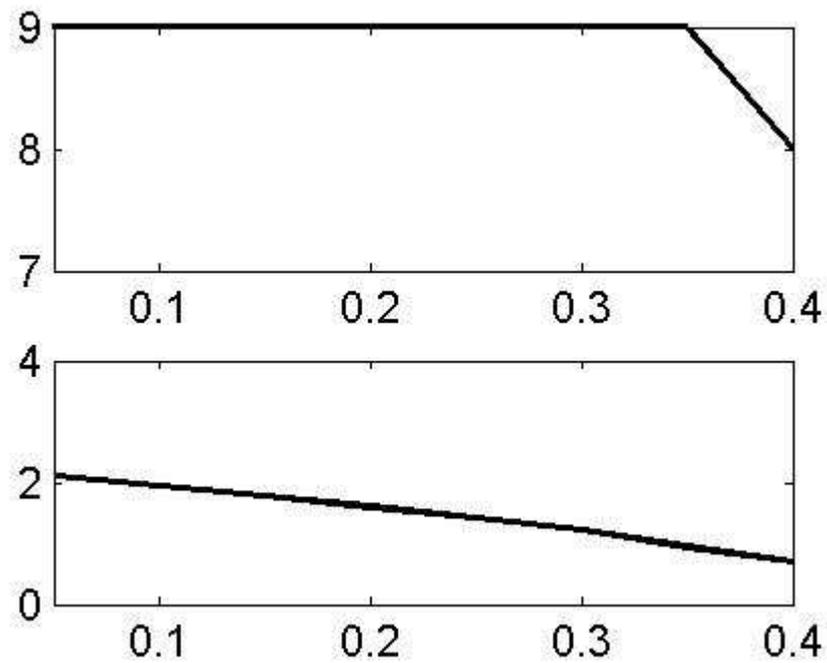

Fig.1. The dependence of $k_c$ (the top panel) and exponent $R$ in the expression (14) (the bottom panel) on the level of the binary noise $a \in [0.05, 0.4]$ (abscissa axis), when the dimension of binary patterns is $N=1000$.

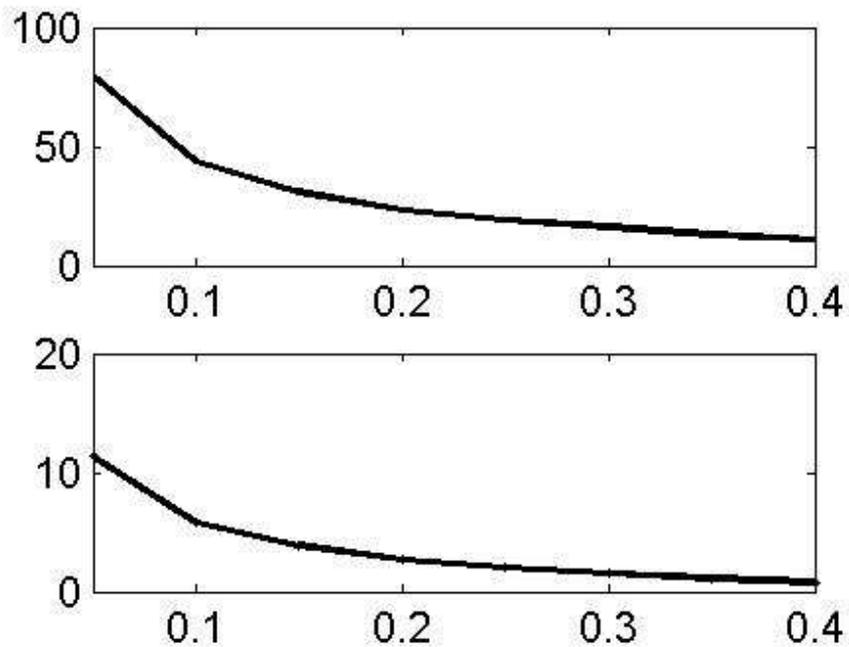

Fig.2. The same as in Fig.1 for the dimension of binary patterns $N=10000$.

### 3.2. *q*-nary identifier

The main idea of the given architecture is based on the following result: when PNN is working far from the limits of its possibility (for example, when the number of patterns is an order of magnitude less than the critical value $M_c$), the retrieval of the correct value of each coordinate occurs at once, during the first processing of this coordinate by the network. (This is because the retrieval error probability falls down exponentially when the number of patterns decreases - see Eqs. (4) and (8)). In other words, in our example to retrieve the pattern correctly the network runs over all the coordinates only once. Then the following approach can be realized.

Suppose we have *M* patterns each of which is described by a set of *N* *q*-nary coordinates. As it is done for PNN3, we represent these coordinates by unit vectors of *q*-dimensional space $\mathbf{R}^q$ (see the expressions (6)). We would like to emphasize, for us substantial meaning of the coordinates are of no importance. It is important only that coordinates can take *q* different values, and in the framework of the vector formalism they are described by unit vectors $\mathbf{e}_l \in \mathbf{R}^q$:

$$X^{(\mu)} = \left(\mathbf{x}_1^{(\mu)}, \mathbf{x}_2^{(\mu)}, \ldots, \mathbf{x}_N^{(\mu)}\right), \quad \mathbf{x}_i^{(\mu)} \in \{\mathbf{e}_l\}_1^q, \quad \mu = 1, \ldots, M. \tag{15}$$

We number all the patterns and write down their numbers in *q*-nary number system. To do this we need $n = \log_q M$ *q*-nary positions. Let us treat these *n* *q*-nary numbers as additional - *enumerated* – coordinates of patterns. Then we add them to the description of each pattern as first *n* coordinates. The enumerated coordinate takes one of *q* possible values, and from this point of view it does not differ from the true coordinates $\mathbf{x}_i^{(\mu)}$ of the pattern. When we pass to the vector formalism, we also represent the values of these enumerated coordinates by *q*-dimensional unit vectors $\mathbf{y}_j \in \{\mathbf{e}_l\}_1^q$. If previously to represent a pattern we needed *N* *q*-nary coordinates, now it is represented by *n+N* such coordinates. In place of the expression (15) we use expanded description of the patterns,

$$\hat{X}^{(\mu)} = \left(\underbrace{\mathbf{y}_1^{(\mu)}, \ldots, \mathbf{y}_n^{(\mu)}}_{\text{enumerated coordinates}}, \underbrace{\mathbf{x}_1^{(\mu)}, \mathbf{x}_2^{(\mu)}, \ldots, \mathbf{x}_N^{(\mu)}}_{\text{true coordinates}}\right), \quad \mathbf{x}_i^{(\mu)}, \mathbf{y}_j^{(\mu)} \in \{\mathbf{e}_l\}_1^q, \mu = 1, \ldots, M.$$

We use these *M* patterns $\hat{X}^{(\mu)}$ of the dimensionality *n+N* to construct PNN-like architecture, whose interconnection matrix, generally speaking, consists of $(n+N)^2$ $(q \times q)$-matrices. However, we suppose that only interconnections $\hat{\mathbf{T}}_{ij}$ between true and enumerated coordinates are nonzero. Other interconnections we forcedly replace by zero (we break these interconnections). Assuming

PNN3 as a basis architecture, we use formula (7) for $i \in [1, n]$ and $j \in [n+1, N+n]$ when calculating the nonzero interconnection matrices,

$$\hat{\mathbf{T}}_{ij} = \begin{cases} \mathbf{T}_{ij} \text{ from Eq.(7)}, & i \leq n, \ j > n, \\ \mathbf{0}, & \text{in the rest cases,} \end{cases} \quad i, j \in [1, N+n]. \qquad (16)$$

The principle of work of such a network is as follows. Suppose we need to recognize distortion of what a pattern $X^{(\mu)}$ (15) is $N$-dimensional $q$-nary image $X$. We transform $N$-dimensional image $X$ into $(n+N)$-dimensional $\hat{X}$, using some *arbitrary* integers from the interval $[1,q]$ in place of the first $n$ enumerated coordinates (in what follows we show that it is not significant what $n$ integers namely are used). In terms of the vector formalism this means that from the image $X = (\mathbf{x}_1, \mathbf{x}_2, ..., \mathbf{x}_N)$ we pass to $\hat{X} = (\mathbf{y}_1, \mathbf{y}_2, ..., \mathbf{y}_n, \mathbf{x}_1, \mathbf{x}_2, ..., \mathbf{x}_N)$.

Just the image $\hat{X}$ of the extended dimensionality $n+N$ is presented for recognition at the input of PNN-like architecture (16). From the construction of the network it is seen that when working the network retrieves the enumerated coordinates only. Consequently, if we work far from the limits of PNN3 possibilities, the correct value of each enumerated coordinate is obtained at once during the first processing (see above). No additional runs over all the coordinates of the pattern will be necessary. So, the number of the pattern, whose distorted copy is the image $X$, will be retrieve at once during one run over $n$ enumerated coordinates. But if the number of the pattern is known, it is not necessary to retrieve correct values of its $N$ true coordinates $\mathbf{x}_i^{(\mu)}$. We simply extract this pattern from the database.

Here the main questions are: How much is the number $n$ of the enumerated coordinates? How large is the additional distortion when we pass from the $N$-dimensional image $X$ to the $(n+N)$-dimensional image $\hat{X}$? It turns out that the number $n$ of the enumerated coordinates is rather small. We remind that $n = \log_q M$. If as $M$ we take the storage capacity of PNN3 given by the expression (9), it is easy to obtain the asymptotic estimate for $n$:

$$n \sim 2 + \frac{\ln N}{\ln q}, \quad N \gg 1.$$

For example, if the number of pixels of the screen is $N \sim 10^4 - 10^6$ and the number of the colors is $q \sim 10^2$, we obtain: $n \approx 4 - 5$. In other words, the number of the enumerated coordinates is negligible small comparing with the number of the true coordinates $N$. So, the additional distortion that appears due to using of arbitrary integers for "numbering" of input image $X$ is negligible small.

In this case the network ceases to be auto-associative one. It works like a *q*-nary perceptron whose interconnections are arranged according the generalized Hebb rule (7). The procedure of recognition of the pattern itself (the retrieval of the correct values of its true coordinates) is absent. Instead of it identification of the number of the pattern takes place. That is why this architecture was called *q-nary identifier*.

The advantages of the *q*-nary identifier are evident. First, the process of recognition speeds up significantly – the network has to retrieve correct values of $n \sim 4-5$ enumerated coordinates only, instead of doing the same for $N \sim 10^4 - 10^6$ coordinates. Second, the computer memory necessary to store interconnections $\mathbf{T}_{ij}$ decrease substantially: now we need only *Nn* such matrices, but not $(n+N)^2$. Though, of course, the storage capacity of such a network somewhat less than the analogous characteristic for PNN3 (at least because we have to work not at the limit of possibilities of PNN3).

In conclusion it is necessary to say that, in fact, up to now associative neural networks were not used for solving complicated practical problems. Seemingly, it is because of the poor recognition characteristics of HM. From this point of view PNNs are very fruitful and promising architectures. We hope that the development of PNNs and other architectures based on PNNs, will lead to increase of interest to the theory of associative neural networks, as well as their practical application.

The work was supported by the program "Intellectual Computer Systems" (the project 2.45) and in part by Russian Basic Research Foundation (grant 06-01-00109).